\documentclass[conference,a4paper]{ieeetran}

\usepackage[latin1]{inputenc}
\usepackage{subeqnarray}
\usepackage{amsfonts}
\usepackage{amsmath}
\usepackage{amssymb}
\usepackage{graphicx}
\usepackage{color}
\usepackage{eurosym}
\usepackage{array,tabularx}

\newenvironment{conditions*}
  {\par\vspace{\abovedisplayskip}\noindent
   \tabularx{\columnwidth}{>{$}l<{$} @{\ : } >{\raggedright\arraybackslash}X}}
  {\endtabularx\par\vspace{\belowdisplayskip}}  
  
\title{\textbf{\Large Embedded Line Scan Image Sensors:\\[-1.5ex] The Low Cost Alternative for High Speed Imaging }\\[0.2ex]}

\author{\authorblockN{ Stef Van Wolputte\authorrefmark{1}, Wim Abbeloos\authorrefmark{1}, Stijn Helsen\authorrefmark{2}, Abdellatif Bey-Temsamani \authorrefmark{2} and Toon Goedem\'e\authorrefmark{1} }
\authorblockA{\authorrefmark{1}EAVISE, Campus De Nayer, KU Leuven, Belgium\\
e-mail: stef.vanwolputte@kuleuven.be, wim.abbeloos@kuleuven.be, toon.goedeme@kuleuven.be\\ 
\authorrefmark{2}Flanders Make, Leuven, Belgium\\
stijn.helsen@flandersmake.be, abdellatif.bey-temsamani@flandersmake.be }
}

\begin{document}
\maketitle
\begin{abstract}
In this paper we propose a low-cost high-speed imaging line scan system. We replace an expensive industrial line scan camera and illumination with a custom-built set-up of cheap off-the-shelf components, yielding a measurement system with comparative quality while costing about 20 times less. We use a low-cost linear (1D) image sensor, cheap optics including a LED-based or LASER-based lighting and an embedded platform to process the images. A step-by-step method to design such a custom high speed imaging system and select proper components is proposed. Simulations allowing to predict the final image quality to be obtained by the set-up has been developed. Finally, we applied our method in a lab, closely representing the real-life cases. Our results shows that our simulations are very accurate and that our low-cost line scan set-up acquired image quality compared to the high-end commercial vision system, for a fraction of the price.

\end{abstract}


\section{Introduction} \label{Introduction}
\begin{figure}
    \begin{center}
    \resizebox{7cm}{!}{\includegraphics{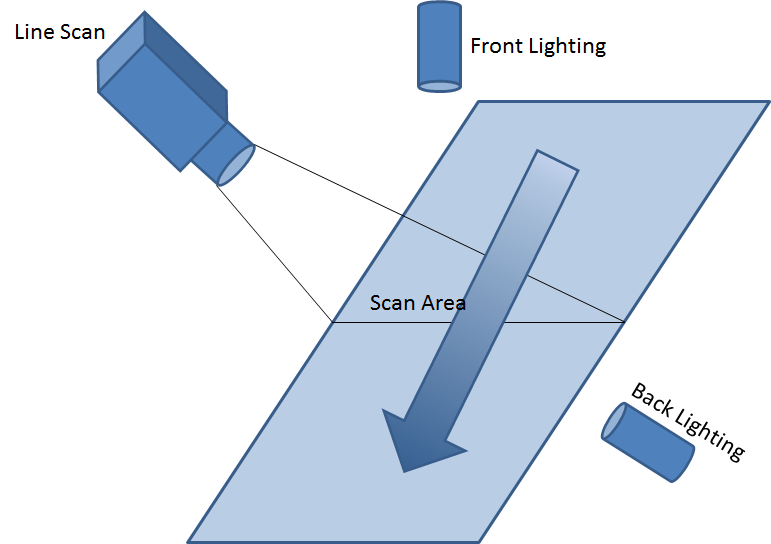}}
\end{center}
\caption{Conceptual overview of a line scan camera set-up.}
\label{TotBlokschema}
\end{figure}

Good sensors form the crux of each industrial process control system.  Vision sensors are often an attractive solution as they do not require physical contact, hence they do not disturb or damage the inspected product.  Some inspections can't be realized using a regular area sensor. Inspecting continuously passing materials or liquids, scanning round objects or very fast moving objects are examples of situations that are very difficult to measure with an area sensor because of its limited frame rate. In these cases a line scan camera is a better option. The sample rates of a line scan sensor can be very high. Thanks to the large rectangular pixels, these cameras are very light-sensitive, such that the integration time can be shorter. Of course the amount of data of each frame is far less compared with a area sensor. Another limitation of area sensors is that they are limited to a linear field of view. While the high frame rate of line scan systems yield a solution to that by applying a \emph{push broom} principle, taking images while the object or the camera moves with the scan line perpendicularly to the motion. A line scan camera set-up is the same compared to a normal camera set-up. Figure~\ref{TotBlokschema} illustrates configurations of the vision set-ups using back and front illumination. These configurations will respectively be used in the set-ups that will be described later on in this paper.

In most industrial applications, relatively expensive high-speed line scan cameras are used, together with specific optics and illumination, such as e.g. telecentric lighting. Our objective is to replace such an industrial line scan vision set-up with a custom made set-up while minimizing the total cost. This cost can even be further decreased if we move from engineering prototype to high quantity production. Of course, this custom set-up must produce the same image quality as the industrial measurement. It is not the goal to compete with the specifications of industrial cameras, the focus lies on the design of a vision system at the lowest possible price for the specific needs of a particular customer. This paper gives a step-by-step guideline to design and select the sensor, lighting and optics for such a set-up. Moreover, for many applications, the image processing algorithms are simple enough to be ported on embedded hardware. Our goal is to propose solutions that include real-time image processing on an embedded platform interfacing directly with the line scan sensor.

\begin{figure}
\begin{center}
    \includegraphics[height=4cm]{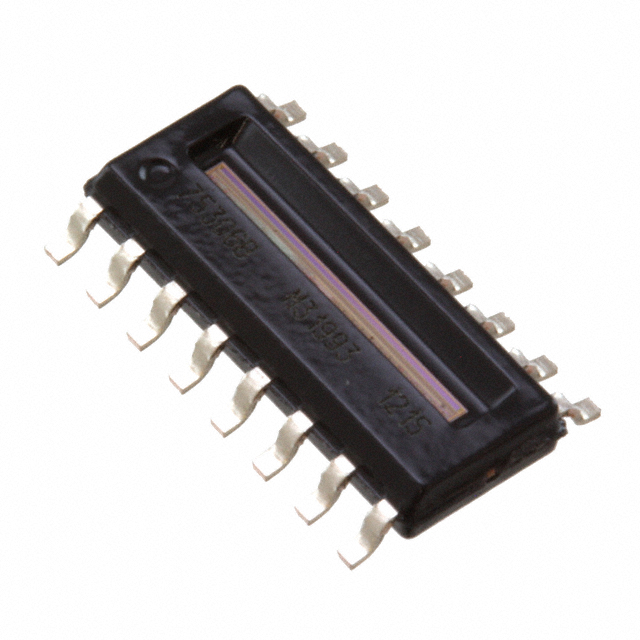}
\end{center}
\caption{The Melexis MLX75306, a 142 pixel linear CMOS sensor.}
\label{sensor}
\end{figure}

\begin{figure}
\begin{center}
    \includegraphics[height=4cm]{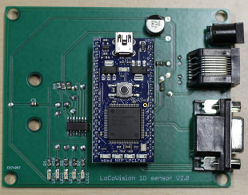}
\end{center}
\caption{Our low-cost line scan PCB. (sensor mounted on the back)}
\label{PCB}
\end{figure}

Recently, an application of high-speed line scan camera for string vibration measurements was presented in \cite{ISMA2014}. This application closely resembles one of the industrial applications we built a set-up for, as explained in section~\ref{HdfApplications}.  In that paper, the authors made use of two industrial Teledyne Dalsa Piranha2 cameras, costing about \euro 2600 each, clearly in outright contrast with the cost of our set-up which is in the order of \euro 100. 
Another example \cite{jasper}, provides a solution for position tracking inside a SEM (scanning electron microscope). A line scan camera with a rate of 1kHz was achieved. Our system can outperform this speed, while using a less complex processing unit.
As mentioned before, our aim is to do the processing of the incoming images locally on the embedded system, such that no external computer has to be used. 
In \cite{konolige}, Konolige et al. showed a low-cost laser distance sensor designed for mapping, localization and obstacle avoidance for mobile robots. The authors used a global shutter 2D CMOS sensor, selecting out only a ROI of 10 rows. Still their major speed limitation is the read out time of the sensor. The use of a linear sensor array, as described in this paper, could offer a faster solution.   

This paper is structured as follows. Firstly, we will go over the different components of an industrial line scan set-up as illustrated in Fig.~\ref{TotBlokschema}, and find low-cost alternatives for each component. Section~\ref{HfdSensor} describes how to select a line scan sensor in order to replace an industrial line scan camera, while section~\ref{HfdOptics} details the choices for low-cost illumination and optical components and section~\ref{HfdPlatform} elaborates on the choice of the embedded processing platform. Then, in section~\ref{HdfApplications}, we demonstrate this approach on two set-ups closely representing two industrial applications: steel wire inspection and distance measurement. Section~\ref{Conclusion} summaries our conclusions.

\section{1D line scan sensor} \label{HfdSensor}
In this section, we show how an industrial line scan camera can be replaced with a custom-made embedded system, based on a linear array sensor. When designing a line scan set-up, all the components have to be matched. We propose to first select a linear sensor array and base the other components selection on this choice.

\paragraph{Pixel resolution}
The first step of a sensor selection should be a mechanical study of the application at hand. With information of the size, motion and expected behavior of the object to be inspected, space limitations, and required measurement accuracy, both the pixel resolution and the frame rate necessary can be computed. 

Based on the study of the application, the field-of-view that is necessary is first identified. Combining this with the measurement accuracy requirement, the resolution of the sensor or the amount of pixels can be calculated. Our experiments show that in normal conditions a subpixel measurement accuracy of a factor 10 can be achieved, further reducing the needed amount of pixels on the sensor by that same number. Linear CMOS sensors of a few tens to thousands of pixels are available on the market, at a very low cost as compared to industrial line scan cameras.  

\paragraph{Frame rate}
One important opportunity of line scan sensors as compared to classical 2D area cameras is that the pixel resolution in transversal and longitudinal directions are independent of each other. The latter can be controlled by the line scan speed, of course taking into account the speed of the object passing by.  We see that even the cheapest sensors on the market already have line speeds of several thousands lines per second (i.e. in the kHz range). This generally is enough for most industrial high speed imaging applications.

With these calculations, minimal numbers for resolution and frame rate of the sensor to be selected can be obtained. For selecting the best suited sensor, obviously also parameters like price, (long term) availability and interface should also be taken into account. Nowadays there are line scan sensors which have a digital output, allowing easy interfacing and therefore reducing the development time.  All these considerations should allow to make a short-list of suited sensors. Basic characteristics of the sensor, like the minimal amount of pixels and line speed, can already filter most sensors which fail or exceed the specifications.

\paragraph{EMVA1288}
The basic calculations described above lead to a short-list of possible sensor candidates. However, the selection of the most suitable sensor is a hard task. Data sheets are mostly incomplete and different suppliers report different units. The EMVA1288 standard \cite{EMVA1288} shows a way to solve this problem and make an unbiased comparison of different vision sensors based on responsivity, light sensitivity, spectral sensitivity, noise, uniformity and linearity.

\section{Illumination and optics} \label{HfdOptics}
After selecting the linear sensor array, a very important part of a line scan imaging set-up is the illumination and optics. Surprisingly, depending on the application, one can even opt to use no lens at all. Indeed, when the mechanical study shows that the size of the sensor is larger than the target object and its variance, a lensless set-up can be chosen to reduce the cost. Of course, this choice will have consequences for the illumination. Both lens-based and lensless set-ups have their pro's and con's, further elaborated on in this section. 

\subsection{Imaging with lens}
\begin{figure}
    \begin{center}
    \includegraphics[height=5cm]{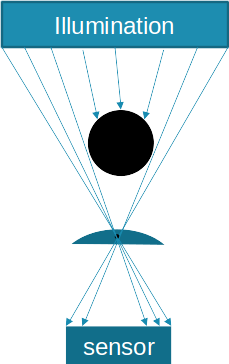} \hspace{1cm} \includegraphics[height=5cm]{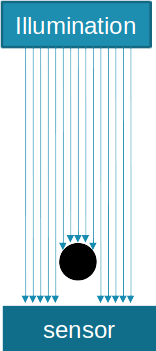}
\end{center}
\caption{Different illumination and optics choices: Left: Imaging with lens and diffuse illumination, Right: Lensless imaging with telecentric illumination.}
\label{lens_lensless}
\end{figure}

A classic optical setup is obtained by placing a lens in front of the sensor, at a distance corresponding to the focal length of the lens. This set-up boils down to a central projection, approximating the pinhole model. Because of the resulting conical field of view, such a lens will allow to measure objects larger than the sensor size. For the illumination, both back and front lighting are possible. When choosing backlighting, such as illustrated at the left of Fig.~\ref{lens_lensless}, one has to take care of the fact that the light source must be at least as large as the field of view and must be ideally diffuse to achieve even illumination. 

Of course, there is a Limited Depth Of Field (DOF) at high magnification ($\approx$1). Also, the object appears larger when close to the sensor and smaller when further away with a linear relation, which can be a problem for the inspection of moving objects. An iris allows to control the amount of light that reaches the sensor, independently of the light source or exposure time. We made a simulation for this set-up, which is available on-line\footnote{http://www.eavise.be/LoCoVision}.

\subsection{Imaging without lens}\label{lensless}
If the mechanical study shows that a lensless set-up is possible, the cost can be reduced by leaving out the lens. This set-up is illustrated in Fig.~\ref{lens_lensless}, on the right. With the light behind the object, a shadow is cast on the sensor surface corresponding to the object's shape.  In order to use such a lensless set-up, the sensor must minimally be as large as the object, and its variance, to be imaged. Also, the back-light illumination must have high degree of telecentricity to achieve a sharp image and to make the object size independent of the object position. 

We made a Matlab simulation which calculates the output image for a lensless set-up with back-lighting, available online~\footnotemark. Using this, the result of an imperfect telecentric light system can be predicted. 

\begin{figure}
    \begin{center}
    \resizebox{10cm}{!}{\includegraphics{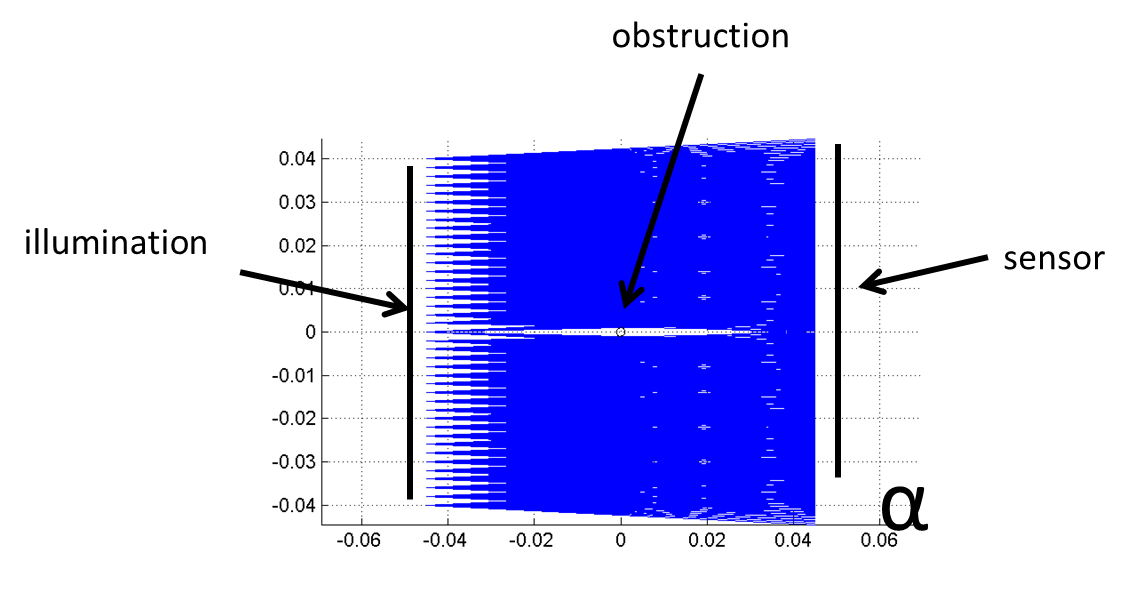}}
\end{center}
\caption{Influence of telecentric slope parameter alfa on light rays paths emerging from a light source.}
\label{rayspaths}
\end{figure}

It is important to note here that even a high-end telecentric light is by defauly imperfect. This is due to the telecentric slope which should be as small as possible (in mrad). However, this parameter is often not mentioned in the datasheet of the telecentric illuminator. We developed a simulation which allows to assess this parameter by modeling its influence  on light rays paths. An example is illustrated in Fig.~\ref{rayspaths}.

When a lensless set-up can be used, the cost of a lens is avoided. However, an important consequence is that the illumination should be telecentric, which can be very expensive. The price of industry-grade telecentric illuminators can rise quickly to the range of \euro 1000. For the low-cost solution we envision in this paper, we worked out two alternative techniques achieving (quasi-)telecentric illumination at a very low-cost, to be detailed in the next subsections: a LED light combined with a parabolic reflector, and a laser diode with a beam expander.

\subsubsection{LED}~\\
For the first technique, we approximated telecentricity by placing a LED, with a parabolic reflector, at a relative large distance from the sensor. The set-up is shown in the top of Fig.~\ref{beam_blok}.

To select an appropriate LED, the first step is to look at the sensor's spectral sensitivity curve, and choose a wavelength with has sufficiently high Quantum Efficiency. We also calculate the required power of the LED.  The electrical energy $E_{e}$ needed to saturate the sensor can be calculated from the light energy needed to saturate the sensor $E_s$, the efficiency of the optics $\eta_{o}$, and the efficiency of the light source $\eta_{led}$ . For a given illuminated surface area, $A$, the electrical energy of the illumination can be found with:

\begin{equation}
    \label{eq:light}
    \begin{split}
		E_e =  \frac{E_{s} A}{\eta_{led}  \eta_{o}}
	\end{split}
\end{equation}

Note that, to achieve good telecentricity, with parallel light beams, the light source should be at a large distance from the sensor. This means the illuminated surface area, $A$, becomes large, and a high power LED is needed.

\begin{figure}
    \begin{center}
    \resizebox{10cm}{!}{\includegraphics{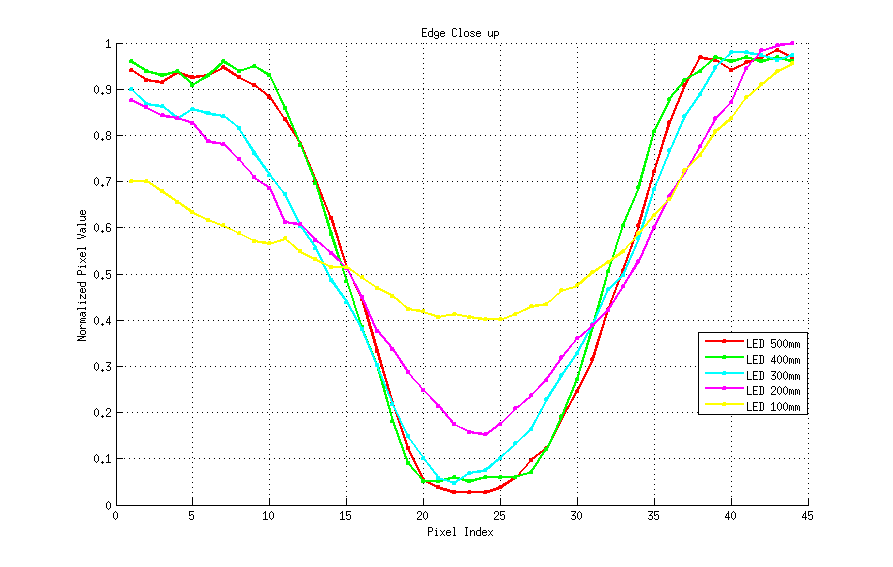}}
\end{center}
\caption{Line images with telecentric lighting with LED placed on different distances from the sensor.}
\label{light_LED}
\end{figure}

To check the quality of the illumination, we placed an object between the camera and the lighting and checked the steepness of the slope in the line scan image values. If the light was perfectly telecentric, the edges of the object would be perfectly sharp. Due to imperfections of the lighting, the slope appears more gradually. 
Figure~\ref{light_LED} shows that the degree of telecentricity clearly increases when the LED is placed on further distances from the sensor. When the LED is placed too close to the sensor, e.g. 100 mm, the yellow line shows that the light is far form telecentric. When the LED is placed further away, the light becomes more telecentric, making the edges sharper.

\begin{figure}
    \begin{center}
    \resizebox{10cm}{!}{\includegraphics{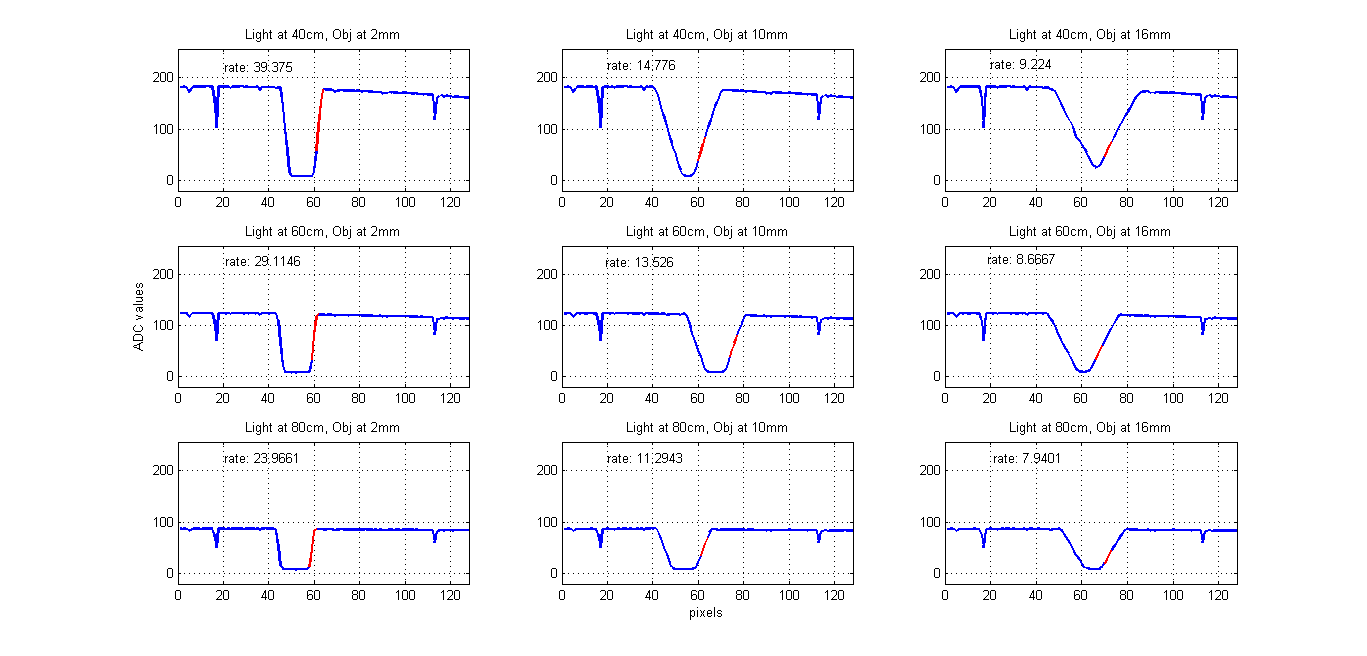}}
\end{center}
\caption{Influence of sensor-object distance on image quality.}
\label{dist_obj_sens}
\end{figure}

We also investigated the influence of distance between object and sensor on image quality. The experimental results are shown in Fig. \ref{dist_obj_sens}.

As clearly shown from Fig. \ref{dist_obj_sens}, the sharpness of the image degrades when the distance between the sensor and the object increases. Therefore as a guideline, the sensor in a lens less set-up should be placed as close as possible to the object to be imaged. In order to quantify the sharpness of the image, a rate parameter is calculated as a light intensity (in ADC values) divided by the number of pixels in the red line of the curve. For every line in the figure (same light-sensor distance), this rate decreases from left to right (increasing object-sensor distance) indicating a degradation of the image sharpness.

\subsubsection{Laser}
\begin{figure}
    \begin{center}
    \resizebox{7cm}{!}{\includegraphics{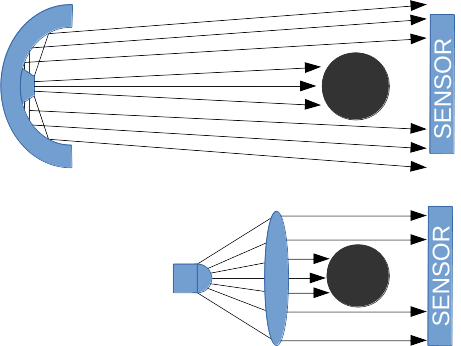}}
\end{center}
\caption{Two systems to obtain telecentric light. Top: LED and reflector.
Bottom: Laser diode with beam expander.}
\label{beam_blok}
\end{figure}

~\\
The previous experiment showed that even when the LED is placed at 500mm, we still get an edge thickness of approximately eight pixels. To achieve a higher degree of telecentricity, and at smaller distances, we tested a set-up using a laser diode. To illuminate the whole sensor, we collimate the diverging light of a TO-CAN laser diode.
This is done using a single planoconvex lens placed in front of the laser diode, as illustrated in the bottom image of Fig.~\ref{beam_blok}.

\begin{figure}
    \begin{center}
    \resizebox{9cm}{!}{\includegraphics{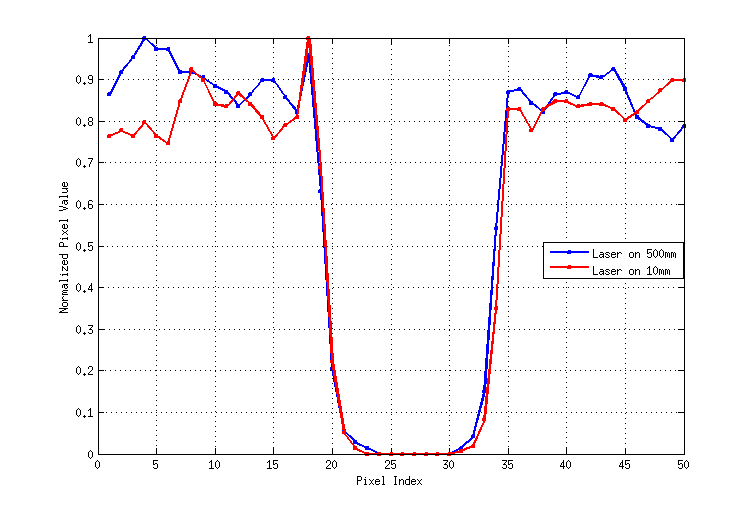}}
\end{center}
\caption{Comparison of laser beam expander placed on 500mm and 10mm from sensor.}
\label{beam500_10}
\end{figure}

\begin{figure}
\begin{center}
   \includegraphics[height=7cm]{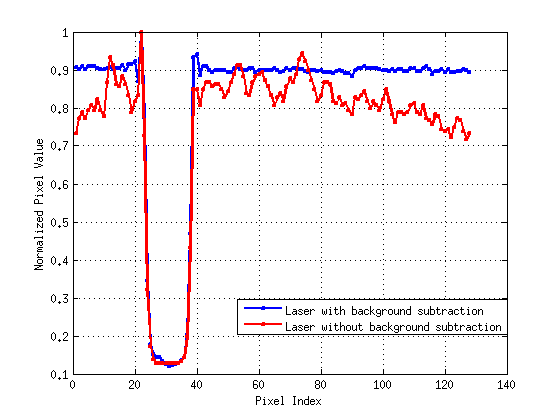}
\end{center}
\caption{Line images with telecentric lighting with a laser without and with background subtraction.}
\label{laser_edge}
\end{figure}

Using the laser with beam expander, we obtain a very sharp image with a one-pixel edge, see Fig.~\ref{beam500_10}. Another benefit is that the distance between sensor and illumination can be greatly reduced. Figure~\ref{beam500_10} also shows that the edge quality does not depend on the distance of the light source.

However, the laser illumination has an important downside: it suffers from laser speckle. This is a random intensity pattern, caused by interference of the monochromatic light waves.  This effect is shown in Fig.~\ref{laser_edge} (red line). Because the laser speckle pattern is constant, we can compensate for it using background subtraction.  In Fig.~\ref{laser_edge}, it is clear that this makes the measurements (blue line) more consistent.

\section{Embedded platform} \label{HfdPlatform}
The last part of the vision system to be selected is a processing unit. Of course, the used algorithm is the major deciding factor. But the sensor interface is also an important parameter. Currently both analog and digital sensors are available. 

\paragraph{Analog sensors}
Most sensors have an analogue output, an external ADC or a processor with on-board ADC is therefore needed. Typically, a DSP is used in this case. An advantage of analogue sensors is that the pixel values can be read out during the next integration time. 

\paragraph{Digital sensors}
When using a sensor with digital output, a much simpler and cheaper processing unit can be used when the algorithms are not too complex. Also the development time will decrease. Through a SPI interface the sensor can be read-out very easily. With the sensor we used it was necessary to wait until the integration time was over before reading out the pixel, causing a delay each period. We tried to minimize this delay by executing our calculations during this integration time.

\section{Applications} \label{HdfApplications}
In the previous chapter we summed up the most important considerations when designing an embedded line scan vision set-up. In this chapter we report on two applications for which we applied this step-by-step guideline. Both applications are based on real-life industrial high speed imaging applications for which a low-cost solution is necessary in order to make it economically feasible. One application is on inspecting steel wire during production, while the objective of the second application is to measure the distance between our sensor and a given surface, while the sensor is placed on a moving object.

\subsection{Wire diameter measurement} 
In steel  production machines, steel wire is produced by twisting together multiple filaments. While the wire is passing by the sensor at very high speeds, the diameter of this composed wire is measured in order to detect fabrication flaws. Moreover, the embedded camera system is intended to perform image processing, trigger the lighting system and raise an alarm when deviations from nominal diameter are detected.

The mechanical study showed that the object diameter could vary between 2.5 and 4 mm. Also, due to standing wave vibrations of the suspended wire, the object has a sinusoidal position variation with an amplitude of maximum 1 mm. The wire moves at a speed of 50 m/s, perpendicularly to the sensor line. The diameter of the wire needs to be measured with a desired accuracy of 50 $\mu$m. The system must be able to detect flaws of a minimal length of 7 mm. 

Based on these findings we computed that the necessary sensor resolution is 120 pixels (even without relying on subpixel calculations) and a line rate of 7.142 kHz. Based on these specifications an other parameter like price, availability and interfacing, we formed a short-list from which we selected the Melexis MLX75306 sensor through the EMVA1288 standard.

This is a 142-pixel CMOS line sensor with a maximal line rate of 9.48 kHz, if 128 pixels are used, and a digital SPI interface. Figure~\ref{sensor} shows a picture of this sensor, which costs about \euro 14 per piece.

Because the object size, and its variance, is smaller than our sensor length (7 mm), it was possible to use a lensless set-up. In order to preform a correct diameter calculation, the backlight would need to be telecentric, which we approximated by the two methods mentioned above, e.g. a LED with a reflector and a laser with a beam expander. To judge the quality of our set-up, we compared it with the set-up that needed to be replaced.

Our sensor was read out by an MBed ARM-based microcontroller. Because we chose a digital linear sensor array, we didn't need any analogue-digital converters. This made it possible to use a lower cost microcontroller. We also implemented all the algorithms on this microcontroller. Every frame, the diameter of the object is calculated. When the diameter exceed its limits, an alarm signal is produced. When controlled lighting is needed, the microcontroller can also trigger the illumination system in a PWM fashion. 

\subsubsection{Simulation}
As mentioned in section~\ref{lensless}, we made a Matlab simulation to predict the inaccuracies due to an imperfect telecentic backlight. After we built our set-up with the LED, we compared the results of the simulation with the practical results.

\begin{figure}
    \begin{center}
    \resizebox{8.5cm}{!}{\includegraphics{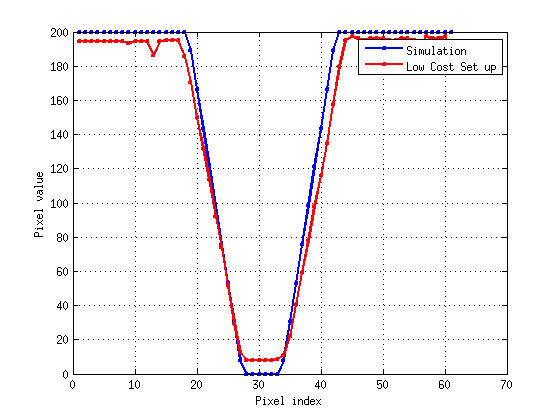}}
\end{center}
\caption{Comparison between the imperfect edges predicted by simulation and real world measurement.}
\label{sim_low}
\end{figure}

We placed a wire between the sensor and a LED, with reflector, placed at 400 mm. We also simulated this exact situation and plotted the slope of the edge from the object. Figure~\ref{sim_low} shows that our simulation (blue) is very similar to the real world measurement (red). 

\subsubsection{Comparison with industrial components}
The goal of this paper was to replace an expensive industrial set-up with a very low cost system, while retaining the same image quality. Therefore, we compared the quality of both systems. 
The industrial set-up, which we needed to replace, consisted out of a Basler Racer camera and an Opto engineering telecentric lighting (LTCLHP023-R).  The total cost of this set-up was \euro 1450.

\begin{figure}
    \begin{center}
    \resizebox{9cm}{!}{\includegraphics{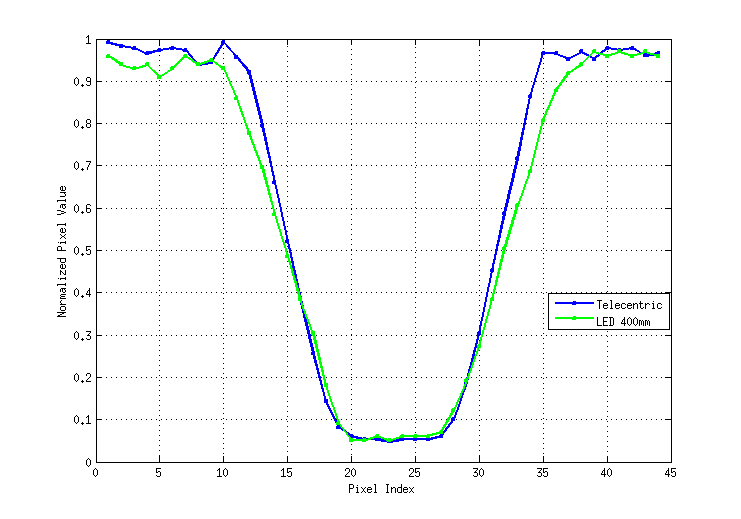}}
\end{center}
\caption{Comparison between industrial and our low-cost illumination.}
\label{tele_400mm}
\end{figure}

First we compared the quality of our LED system with the telecentric illumination. To compare both, we again checked the object edge steepness, using the industrial camera. Figure~\ref{tele_400mm} plots the image intensities while using industrial telecentric lighting and our LED-based alternative. While the object edge with the expensive telecentric lighting is about 8 pixels wide, our low-cost LED lighting doesn't do much worse with 9 pixels of edge width.

\begin{figure}
    \begin{center}
    \resizebox{9cm}{!}{\includegraphics{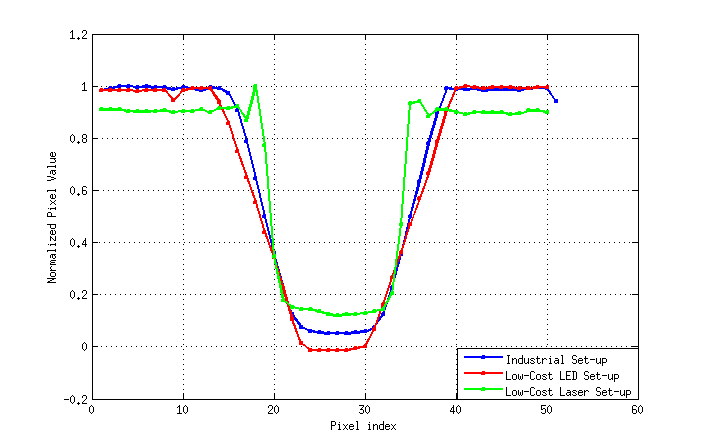}}
\end{center}
\caption{Comparison between high cost camera and telecentric illumination and our low cost set-up, 1D sensor with approximate telecentric illumination.}
\label{low_high}
\end{figure}

Next we tested the quality of our total set-up. Therefore we compared the output of the industrial set-up with our custom made system consisting if the low-cost linear Melexis sensor and LED or laser illumination. 
The steepness of the object slope which we get with this system in given in Fig.~\ref{low_high} by the blue line. When we used our own set-up, with the LED placed on 400 mm, we achieved a slope as given by the red line. One can see in Fig.~\ref{low_high} that the industrial set-up (blue) generated a slightly steeper edge than the LED-based set-up, but the difference is minimal.

To achieve even higher quality images, we implemented the solution based on a laser with beam expander. This allowed to get even steeper slopes compared to the industrial telecentric illumination, shown by the green line in Fig.~\ref{low_high}.

Of course the price of the new developed vision system is much cheaper than the commercial one. The bill of materials of the set-up we designed is given in table~\ref{BOM}. Remark that these are the prices when ordered in low amounts, when ordering higher quantities the price can be further reduced. The price of the casing is not taken into account. The proposed vision system is more than 20 times cheaper than the industrial components-based solution.

\begin{table}
\caption{Pricing total set-up.}\label{BOM}
\centering%
\begin{tabular}{lll}
\hline
\textit{Sensor PCB} & \textit{} & \textit{Price (EUR)} \\
\hline
1) & ARM Cortex-M3 + peripherals & 20\\
2) & Power components  & 5\\
3) & Connectors & 10\\
4) & MLX75306 + peripherals & 16\\
\hline\\
\textit{LED Driver} & \textit{} & \textit{Price (EUR)} \\
\hline
1) & Power LED & 7\\
2) & 4 degree optic reflector  & 4\\
3) & LED holder & 1\\
4) & PCB + driver & 4\\
\hline\\
\textit{Laser Beam expander} & \textit{} & \textit{Price (EUR)} \\
\hline
1) & Laser diode & 2\\
2) & Planoconvex lens  & 10\\
3) & PCB + driver & 3\\
\hline\\
 & Total Price using LED   & 67\\
 & Total Price using Laser & 66\\
 \hline\\
\end{tabular}
\end{table}

\subsection{Distance estimation}
\begin{figure}
    \begin{center}
    \includegraphics[height=5cm]{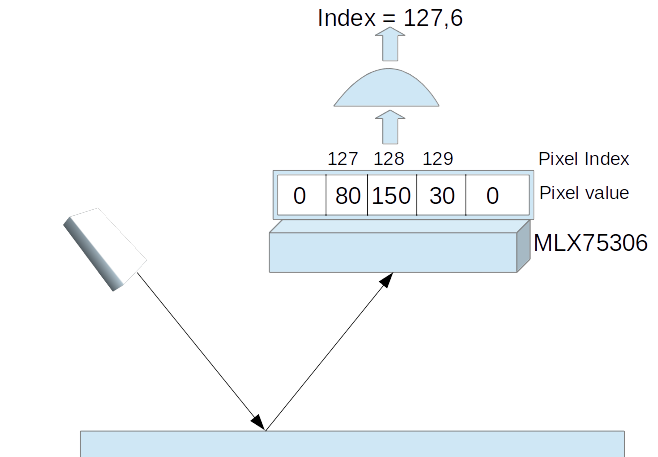}
\end{center}
\caption{Block scheme of the distance estimation set-up.}
\label{tenneco}
\end{figure}

We also applied our method on a second case. The goal here was to measure the distance below the sensor, installed in a moving system. The mechanical requirements of this case are as follows. The distance measurement needs to be accurate to 1 mm over a range of 200 mm. For a first prototype, the speed of the object is limited to 20 m/s while the interval between two samples had to be 7 mm.
Hereby we could use the same sensor as used in the previous application. However we needed to make some modifications to our previous set-up. We use a laser as front-lighting to calculate the height in a \emph{sheet of light} 3D scanning fashion, as illustrated in Fig.~\ref{tenneco}. For the required field-of-view of at least 20 cm it is clear that a lens is needed in this case. 

\begin{figure}
    \begin{center}
    \includegraphics[height=5cm]{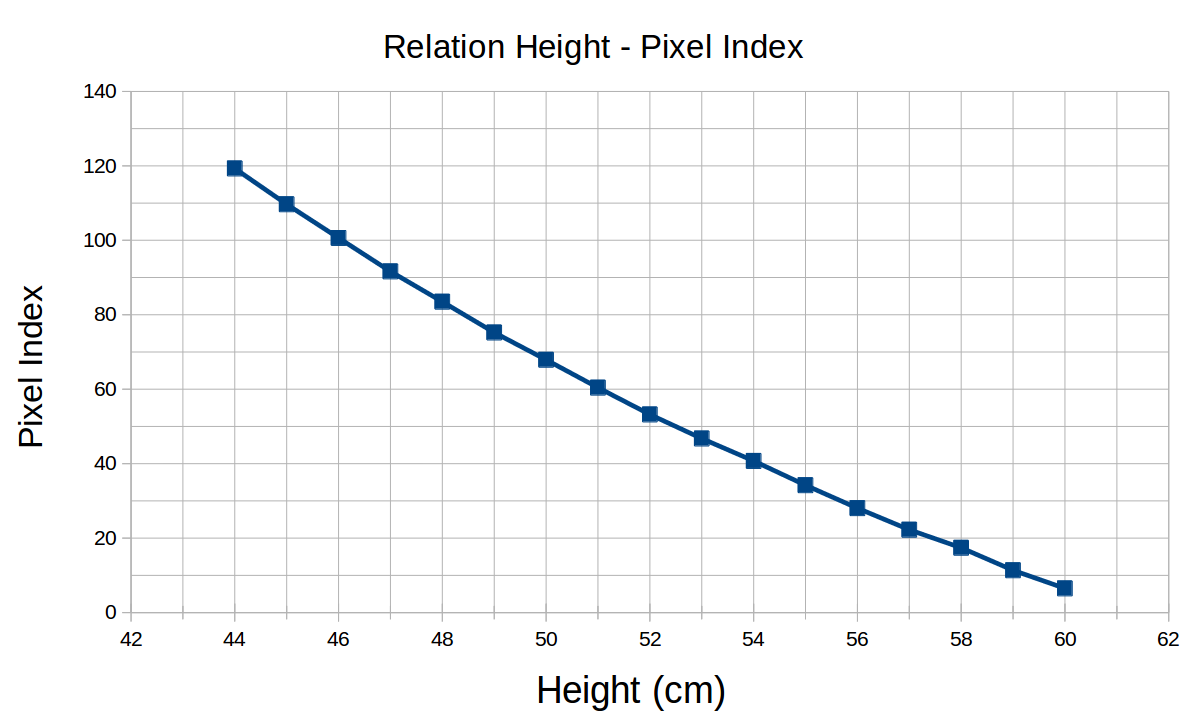}
\end{center}
\caption{Relation between the distance and pixel index.}
\label{height}
\end{figure}

We implemented the total processing and the triangulation algorithm on the embedded system interfacing with the sensor. This way, the output of our camera system module is only the height between the sensor and the surface. To calculate the height, we first search the brightest pixels in the line scan sensor. Then, we reach sub-pixel accuracy by fitting a parabolic function through the three brightest pixels. The subpixel index of the apex of this parabola is then mapped to it corresponding height and outputted. The relation between the height under the sensor and the pixel index is given in Fig.~\ref{height}.

The total system reaches a speed of 3.5 kHz, mainly bounded by the digital interface between sensor and microcontroller. This makes it possible that the studied system moves at a speed of  25 m/s. Because of the use of a lens the total system is a bit more expensive. But with \euro 50 for the optics and \euro 10 for the laser driver, the total cost of \euro 111 is still very low for a total solution. 

\section{Conclusions}
\label{Conclusion}
In this paper we propose a low-cost line scan vision system  for a high speed imaging. We showed that we could achieve a similar image quality in comparison to a high-end industrial vision system for only a fraction of the price.
We presented a guideline which leads to a customized line scan set-up and provides simulations which can predict the image acquired by the set-up. We also tested the developed low-cost vision system on two set-ups closely representing both industrial applications: quality inspection of high-speed copper production and height estimation of a moving system. 

We showed that the developed system reaches all the requirements of these applications. The images acquired by the developed low-cost vision system are comparable in quality to the images acquired using a high-end industrial vision system. However, the cost of the new system is only about \euro 100 which makes it 20 times cheaper than the commercial system. In order to target even higher speed applications, faster image acquisition rate can be achieved with an analog sensor, combined with an embedded platform.

\section*{Acknowledgment}
This work is partially funded by IWT and Flanders Make via the LocoVision ICON project.

\end{document}